\documentclass[vecarrow]{svmult}
\usepackage{mathptmx}
\usepackage{helvet}
\usepackage{courier}
\usepackage{makeidx}
\usepackage{amsmath,amsfonts}
\usepackage{graphicx}
\usepackage{cite}
\usepackage{centercaption}
\usepackage[bottom]{footmisc}
\usepackage{bm}

\begin{document}

\input{opening_Fayet-Wolhart_ARK2020.inp} \abstract{%
This paper revisits a three-loop spatial linkage that was proposed in an ARK
2004 paper by Karl Wohlhart (as extension of a two-loop linkage proposed by
Eddie Baker in 1980) and later analyzed in an ARK 2006 paper by Diez-Mart%
\'{\i}nez et. al. A local analysis shows that this linkage has a finite
degree of freedom (DOF) 3 (and is thus overconstrained) while in its
reference configuration the differential DOF is 5. It is shown that its
configuration space is locally a smooth manifold so that the reference
configuration is not a c-space singularity. It is shown that the
differential DOF is locally constant, which makes this linkage shaky (so
that the reference configuration is not a singularity). The higher-order
local analysis is facilitated by the computation of the kinematic tangent
cone as well as a local approximation of the c-space.
} \input{keywords.inp}

\section{Introduction}

\begin{figure}[tbp]
\centering%
\includegraphics[width=0.75
\linewidth]{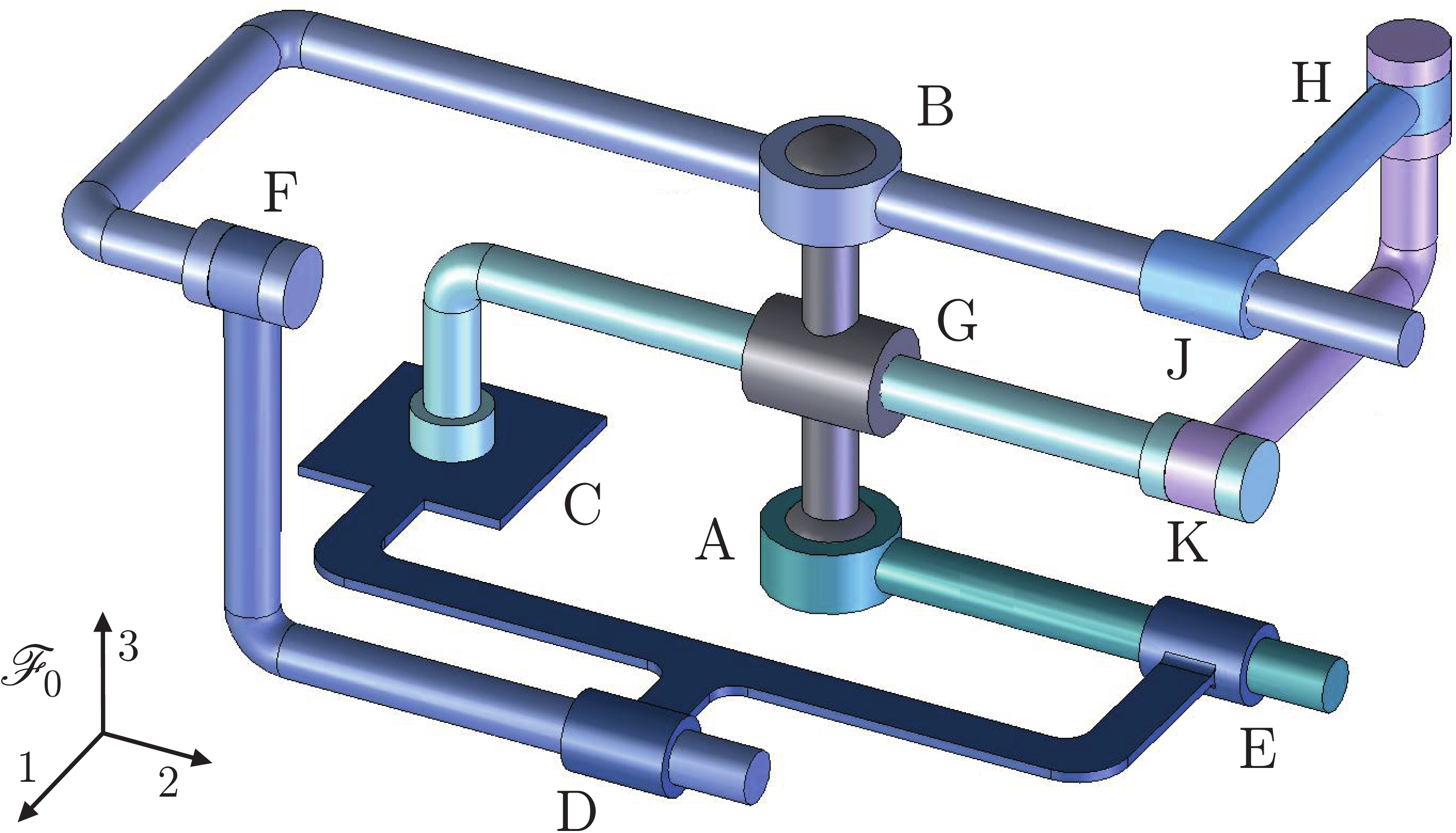}
\caption{Multiloop linkage presented by Fayet (CAD drawing curtesy of Jose
Rico).}
\label{figFayet}
\end{figure}

In \cite{Baker1980}, Baker proposed a two-loop linkage as an example to
demonstrate the application of his method for identification of the relative
twist of any two bodies, and thus their instantaneous relative mobility, in
a mechanism. This linkage was also used by Davies \cite{Davies1981} to
demonstrate his graph representation of the kinematic topology. Fayet \cite%
{FAYET1995_1} extended it to a three-loop linkage and used it as an example
for application of his method to determine the 'relative screw space', i.e.
the subspace of $se\left( 3\right) $ defined by the possible relative
twists. An efficient method for identification of the relative twists was
proposed by Wohlhart \cite{Wohlhart-ARK2004}, and later Diez-Mart\'{\i}nez
et al. \cite{Rico-ARK2006} proposed a method for determination of the finite
relative mobility. In both publications, a variant of Fayet's linkage shown
in Fig. \ref{figFayet} was used as an example, and it was analyzed in the
shown reference configuration. It was shown in \cite{Wohlhart-ARK2004} that
in this reference configuration the differential DOF is 5. In \cite%
{Rico-ARK2006}, the finite DOF was found to be 3. The differential and local
DOF being different may suggest that the reference configuration is a
kinematic singularity. This question has not yet been addressed, and will be
answered in this paper by means of a local mobility analysis. The latter
determines the finite mobility at the reference configuration in Fig. \ref%
{figFayet} as well as the differential mobility when the linkage performs
finite motions through that configuration. It will be shown that this
configuration is a regular point of the c-space, that it is not a kinematic
singularity, and that the differential DOF is always 5, which renders the
linkage shaky. A mechanism is shaky iff, in a regular configuration, its
differential (instantaneous) DOF is higher than its local (finite) DOF. The
analysis makes use of the kinematic tangent cone, which provides the
mathematical framework for such a local analysis as introduced in \cite%
{Lerbet1999,JMR2016,JMR2018,CISMMueller2019}, and a local approximation of
the c-space. The computation is enabled by the closed form and recursive
relations for the higher-order expressions in terms of joint screws \cite%
{Mueller-MMT2019}. A Mathematica$^{\copyright }$ notebook with all
computations for the linkage in Fig. \ref{figFayet} can be obtained from the
author.

\section{Kinematic Topology}

\begin{figure}[tbp]
\centering\includegraphics[width=0.8%
\linewidth]{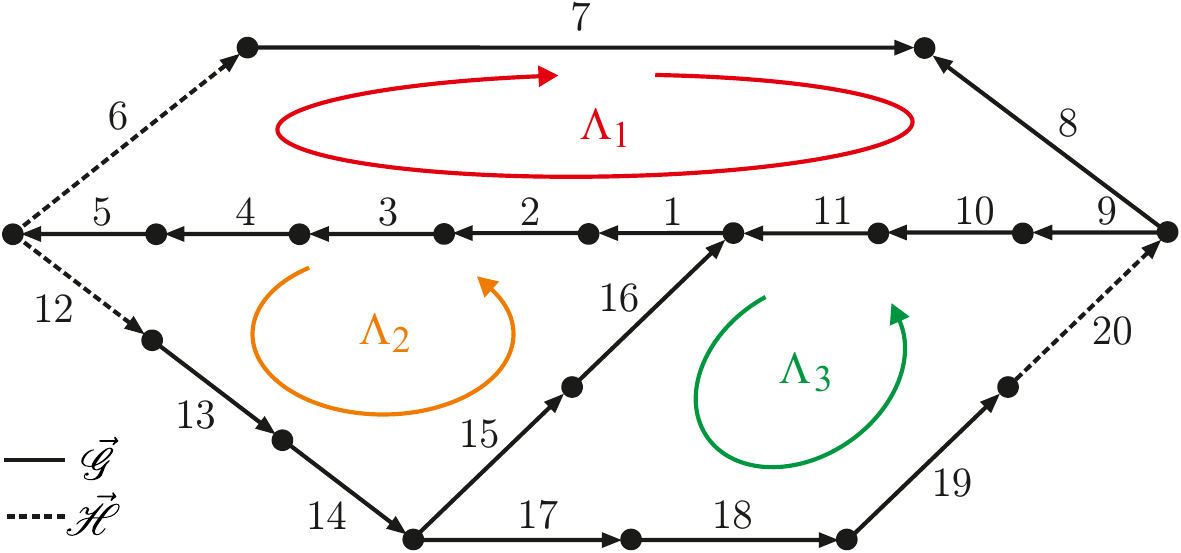}
\caption{Tree $\vec{\mathcal{G}}$, co-tree $\vec{\mathcal{H}}$, and FCs $%
\Lambda _{l}$ of the topological graph $\vec{\Gamma}$ for the linkage in Fig
1.}
\label{figGraph}
\end{figure}

The kinematic topology of a linkage is represented by a non-directed
(multi)graph $\Gamma $ \cite{Davies1981,Robotica2018}. Vertices represent
bodies/links, and edges represent (1-DOF) joints. The linkage in Fig. \ref%
{figFayet} comprises 10 joints (2 spherical, 4 cylindrical, 1 planar, 3
revolute) and 8 links. The cylindrical, spherical, and planar joints are
modeled as combination of helical joints \cite{Wohlhart-ARK2004,Rico-ARK2006}%
. The kinematic model thus consists of $n=20$ helical joints (6 prismatic,
14 revolute) and $N=17$ links. The topological graph possesses $\gamma =$ $%
N-n+1=3$ fundamental cycles (FC), also called fundamental loops, denoted
with $\Lambda _{l},l=1,\ldots ,\gamma $. A co-tree $\mathcal{H}$ can be
introduced consisting of exactly one edge in each of the $\gamma $ FCs.
Eliminating the co-tree edges yields a spanning tree $\mathcal{G}$. For the
kinematics modeling, an associated directed graph $\vec{\Gamma}$ is
introduced. The edge directions indicate the polarity of joint variables.
This defines the corresponding directed tree $\vec{\mathcal{G}}$ and co-tree 
$\vec{\mathcal{H}}$, as shown in Fig. \ref{figGraph} (vertex/body labels are
omitted). FCs induce an ordering according to how edges appear when
traversing a FC. All edges are aligned with the FCs except edges 15 and 16.

The correspondence of edges of $\vec{\Gamma}$ (i.e. revolute/prismatic
joints in the model) to the multi-DOF joints of the linkage is as follows: 
\emph{Joint A}: 1,2,3; \emph{B}: 9,10,11; \emph{C}: 12,13,14; \emph{D}: 6,7; 
\emph{E}: 4,5; \emph{F}: 8; \emph{G}: 15,16; \emph{H}: 18; \emph{J}: 19,20; 
\emph{K}: 17.\emph{\ } 

\section{Kinematic Model}

Edge $i$ of the topological graph corresponds to one of the helical joints,
modeled with a screw coordinate vector, denoted with $\mathbf{Y}%
_{i},i=1,\ldots ,n$, and a joint variable, denoted with $q_{i},i=1,\ldots ,n$%
. The latter are summarized in the vector $\mathbf{q}\in {\mathbb{V}}^{n}={%
\mathbb{R}}^{6}\times {\mathbb{T}}^{14}$. The configuration in Fig. \ref%
{figFayet} is used as zero reference configuration $\mathbf{q}_{0}=\mathbf{0}
$. The screw coordinate vectors are determined in this reference
configuration, and represented in the spatial frame $\mathcal{F}_{0}$ when
located at joint $A$ (adopted from \cite{Rico-ARK2006}):%
\begin{eqnarray*}
\mathbf{Y}_{1} &=&(1,0,0,0,0,0)^{T},\mathbf{Y}_{2}=\mathbf{Y}%
_{5}=(0,1,0,0,0,0)^{T} \\
\mathbf{Y}_{3} &=&\mathbf{Y}_{11}=(0,0,1,0,0,0)^{T},\mathbf{Y}_{4}=\mathbf{Y}%
_{6}=\mathbf{Y}_{13}=\mathbf{Y}_{15}=\mathbf{Y}_{19}=\left(
0,0,0,0,1,0\right) ^{T} \\
\mathbf{Y}_{7} &=&(0,1,0,0,0,1)^{T},\mathbf{Y}_{8}=\left(
0,1,0,-1,0,1\right) ^{T},\mathbf{Y}_{9}=\left( 1,0,0,0,1,0\right) ^{T} \\
\mathbf{Y}_{10} &=&\mathbf{Y}_{20}=\left( 0,1,0,-1,0,0\right) ^{T},\mathbf{Y}%
_{12}=\left( 0,0,0,1,0,0\right) ^{T} \\
\mathbf{Y}_{14} &=&(0,0,1,-1,0,0)^{T},\mathbf{Y}_{16}=\left(
0,1,0,-1/2,0,0\right) ^{T} \\
\mathbf{Y}_{17} &=&\left( 0,1,0,-1/2,1,0\right) ^{T},\mathbf{Y}_{18}=\left(
0,0,1,1,1,0\right) ^{T}.
\end{eqnarray*}%
The instantaneous joint screw coordinates in an arbitrary configuration are
denoted with ${\mathbf{S}_{i}}\left( \mathbf{q}\right) $. They are found
from the joint screw coordinates in the reference configuration by a frame
transformation of $\mathbf{Y}_{i}$ to the current configuration \cite%
{CISMMueller2019,Mueller-MMT2019}. For instance ${\mathbf{S}_{9}}%
\hspace{-0.5ex}%
\left( \mathbf{q}\right) =\mathbf{Ad}_{\exp (\mathbf{Y}_{1}q_{1})\cdot
\ldots \cdot \exp (\mathbf{Y}_{{8}}q_{8})}\mathbf{Y}_{9}$, where $\mathbf{Ad}
$ is the matrix transforming screw coordinates \cite{SeligBook}. Indeed, ${%
\mathbf{S}_{i}}%
\hspace{-0.5ex}%
\left( \mathbf{0}\right) =\mathbf{Y}_{i}$ in the zero reference
configuration.

Denote with $\mathbf{J}_{l}$ the constraint Jacobian in the velocity
constraints $\mathbf{J}_{l}(\mathbf{q})\dot{\mathbf{q}}=\mathbf{0}$ for the
FC $\Lambda _{l},l=1,\ldots ,3$. The overall constraint Jacobian for the $%
\gamma =3$ FCs, according to the orientation of the topological graph, is
the $18\times 20$ matrix%
\begin{equation}
\mathbf{J}=\left( 
\begin{array}{cccccccccccccccccccc}
\mathbf{S}_{1} & \mathbf{S}_{2} & \mathbf{S}_{3} & \mathbf{S}_{4} & \mathbf{S%
}_{5} & \mathbf{S}_{6} & \mathbf{S}_{7} & \mathbf{S}_{8} & \mathbf{S}_{9} & 
\mathbf{S}_{10} & \mathbf{S}_{11} & \mathbf{0} & \mathbf{0} & \mathbf{0} & 
\mathbf{0} & \mathbf{0} & \mathbf{0} & \mathbf{0} & \mathbf{0} & \mathbf{0}
\\ 
\mathbf{S}_{1} & \mathbf{S}_{2} & \mathbf{S}_{3} & \mathbf{S}_{4} & \mathbf{S%
}_{5} & \mathbf{0} & \mathbf{0} & \mathbf{0} & \mathbf{0} & \mathbf{0} & 
\mathbf{0} & \mathbf{S}_{12} & \mathbf{S}_{13} & \mathbf{S}_{14} & \mathbf{S}%
_{15} & \mathbf{S}_{16} & \mathbf{0} & \mathbf{0} & \mathbf{0} & \mathbf{0}
\\ 
\mathbf{0} & \mathbf{0} & \mathbf{0} & \mathbf{0} & \mathbf{0} & \mathbf{0}
& \mathbf{0} & \mathbf{0} & \mathbf{S}_{9} & \mathbf{S}_{10} & \mathbf{S}%
_{11} & \mathbf{0} & \mathbf{0} & \mathbf{0} & -\mathbf{S}_{15} & -\mathbf{S}%
_{16} & \mathbf{S}_{17} & \mathbf{S}_{18} & \mathbf{S}_{19} & \mathbf{S}_{20}%
\end{array}%
\right)  \label{J}
\end{equation}%
where the three block rows correspond to the $\mathbf{J}_{l}$. The system of 
$6\gamma =18$ velocity constraints is thus $\mathbf{J}(\mathbf{q})\dot{%
\mathbf{q}}=\mathbf{0}$. Note that edges 15 and 16 are directed opposite to $%
\Lambda _{3}$.

The geometric loop closure constraints of the FC $\Lambda _{l},l=1,\ldots
,\gamma $ is $f_{l}\left( \mathbf{q}\right) =\mathbf{I}$ with the constraint
mappings expressed by the product of exponentials%
\begin{eqnarray}
f_{1}\left( \mathbf{q}\right) &=&\exp (\mathbf{Y}_{1}q_{1})\cdot \ldots
\cdot \exp (\mathbf{Y}_{{11}}q_{11}) \\
f_{2}\left( \mathbf{q}\right) &=&\exp (\mathbf{Y}_{1}q_{1})\cdot \ldots
\cdot \exp (\mathbf{Y}_{5}q_{5})\exp (\mathbf{Y}_{12}q_{12})\cdot \ldots
\cdot \exp (\mathbf{Y}_{16}q_{16}) \\
f_{3}\left( \mathbf{q}\right) &=&\exp (\mathbf{Y}_{9}q_{9})\exp (\mathbf{Y}%
_{10}q_{10})\exp (\mathbf{Y}_{11}q_{11})\exp (-\mathbf{Y}_{16}q_{16})\exp (-%
\mathbf{Y}_{15}q_{15})  \notag \\
&&\cdot \exp (\mathbf{Y}_{17}q_{17})\exp (\mathbf{Y}_{18}q_{18})\exp (%
\mathbf{Y}_{19}q_{19})\exp (\mathbf{Y}_{20}q_{20})
\end{eqnarray}%
where the ordering is defined by the FCs and the sign in the exponential by
the direction of edges relative to the FCs. The c-space is then the real
analytic variety%
\begin{equation}
V:=\left\{ \mathbf{q}\in {\mathbb{V}}^{20}|f_{l}\left( \mathbf{q}\right) =%
\mathbf{I},l=1,\ldots ,3\right\} .  \label{V}
\end{equation}%
The principle aim of a mobility analysis consists in determining the
dimension (i.e. the DOF) and the topology of $V$ (i.e. the mobility). The
differential DOF at $\mathbf{q}\in V$ is $\delta _{\mathrm{diff}}\left( 
\mathbf{q}\right) =\dim \ker \mathbf{J}\left( \mathbf{q}\right) =n-\mathrm{%
rank}~\mathbf{J}\left( \mathbf{q}\right) $. The differential mobility is
described by vectors $\mathbf{x}\in \ker \mathbf{J}\left( \mathbf{q}\right) $%
. The finite local DOF at $\mathbf{q}\in V$ is the local dimension of $V$,
denoted $\delta _{\mathrm{loc}}\left( \mathbf{q}\right) =\dim _{\mathbf{q}}V$%
.

\emph{C-space singularities} are non-smooth points of $V$ (where $V$ is not
a smooth manifold). \emph{Kinematic singularities} are points where the
differential DOF is not constant in any neighborhood of $\mathbf{q}$ in $V$.
C-space singularities are kinematic singularities. The opposite is not
necessarily true. \emph{Constraint singularities} are points where the
constraint Jacobian is not full rank. C-space and kinematic singularities
are constraint singularities. The opposite is not 
\mbox{necessarily true
(e.g. overconstrained mechanisms).}

\section{Smooth Finite Motions through the Reference Configuration}

A motion of the linkage corresponds to a curve in $V$. At a given
configuration $\mathbf{q}\in V$, the tangents to smooth finite curves in $V$
through $\mathbf{q}$, i.e. to smooth motions, form the kinematic tangent
cone $C_{\mathbf{q}}^{\mathrm{K}}V:=\{\dot{\sigma}|\sigma \in \mathcal{C}_{%
\mathbf{q}}\}\subset {\mathbb{R}}^{n}$, where $\mathcal{C}_{\mathbf{q}}$ is
the class of smooth curves in $V$ through $\mathbf{q}\in V$ \cite%
{CISMMueller2019}. With the parameterization in terms of joint variables it
can be simply regarded as a formal definition of the set of all velocity
vectors $\dot{\mathbf{q}}$ such that $\mathbf{q}\left( t\right) $ is a
smooth curve in $V$ and satisfies time derivative of the loop constraints of
any order. It can thus be defined in terms of higher-order derivatives of
the velocity constraints \cite{Lerbet1999,JMR2016,JMR2018}. Denoting $%
H_{l}^{\left( 1\right) }%
\hspace{-0.5ex}%
(\mathbf{q},\dot{\mathbf{q}}):=\mathbf{J}_{l}(\mathbf{q})\dot{\mathbf{q}}$
and $H_{l}^{\left( i\right) }%
\hspace{-0.5ex}%
(\mathbf{q},\dot{\mathbf{q}},\ldots ,\mathbf{q}^{\left( i\right) }):=\frac{%
d^{i-1}}{dt^{i-1}}H_{l}^{\left( 1\right) }%
\hspace{-0.6ex}%
\left( \mathbf{q},\dot{\mathbf{q}}\right) $, the $i$th-order constraints for
the FC $\Lambda _{l}$ is%
\begin{equation}
H_{l}^{\left( i\right) }%
\hspace{-0.5ex}%
(\mathbf{q},\dot{\mathbf{q}},\ldots ,\mathbf{q}^{\left( i\right) })=\mathbf{0%
}.  \label{ConstrHi}
\end{equation}%
The kinematic tangent cone is then determined as%
\begin{equation}
{C_{\mathbf{q}}^{\text{K}}V}=K_{\mathbf{q}}^{\kappa }\subseteq \ldots
\subseteq K_{\mathbf{q}}^{3}\subseteq K_{\mathbf{q}}^{2}\subseteq {K_{%
\mathbf{q}}^{1}}  \label{CqV}
\end{equation}%
where $K_{\mathbf{q}}^{i}$ is a cone (in $\mathbf{x}$) defined as%
\begin{equation}
\begin{array}{ll}
K_{\mathbf{q}}^{i}:=\{\mathbf{x}|\exists \mathbf{y},\mathbf{z},\ldots \in {%
\mathbb{R}}^{20}: & H_{l}^{\left( 1\right) }%
\hspace{-0.6ex}%
\left( \mathbf{q},\mathbf{x}\right) =\mathbf{0},H_{l}^{\left( 2\right) }%
\hspace{-0.6ex}%
\left( \mathbf{q},\mathbf{x},\mathbf{y}\right) =\mathbf{0},H_{l}^{\left(
3\right) }%
\hspace{-0.6ex}%
\left( \mathbf{q},\mathbf{x},\mathbf{y},\mathbf{z}\right) =\mathbf{0},\ldots
\\ 
& \ldots ,H_{l}^{\left( i\right) }%
\hspace{-0.6ex}%
\left( \mathbf{q},\mathbf{x},\mathbf{y},\mathbf{z,\ldots }\right) =\mathbf{0}%
,l=1,2,3\}.%
\vspace{-1ex}%
\end{array}
\label{Ki}
\end{equation}%
The sequence (\ref{CqV}) terminates with finite $\kappa $, which is specific
to the linkage and depends on the configuration. The crucial point of this
formulation is that the higher-order mappings $H_{l}^{\left( i\right) }$ can
be evaluated in closed form or recursively by means of simple vector
operations \cite{Mueller-MMT2019,CISMMueller2019}. A Mathematica$^{%
\copyright }$ implementation is available at \cite{Mueller-MendeleyData2019}.

The kinematic tangent cone provides a local description of the manifolds
intersecting at a c-space singularity $\mathbf{q}$. It thus allows to
identify motion bifurcations, including non-transversal intersections \cite%
{LopezCustodio-MMT2020}. The calculation shows that the first- and
second-order cone is a 5-dimensional and 3-dimensional vector space,
respectively:%
\begin{equation*}
\begin{array}{ll}
K_{\mathbf{q}_{0}}^{1}=\{\mathbf{x}\in {\mathbb{R}}^{20}| & 
x_{1}=0,x_{2}=u,x_{3}=v,x_{4}=0,x_{5}=w+2v,x_{6}=-v,x_{7}=w/2, \\ 
& x_{8}=r,x_{9}=-v,x_{10}=s,x_{11}=w+2v,x_{12}=-s,x_{13}=-w-u-2v, \\ 
& x_{14}=-w-2v,x_{15}=-r+s,x_{16}=w+2v,x_{17}=v, \\ 
& x_{18}=-w-v-r+s,x_{19}=-w,x_{20}=w;\ r,s,u,v,w\in {\mathbb{R}}\}.%
\end{array}%
\end{equation*}%
\begin{equation*}
\begin{array}{ll}
K_{\mathbf{q}_{0}}^{2}=\{\mathbf{x}\in {\mathbb{R}}^{20}| & 
x_{1}=0,x_{2}=-u,x_{3}=0,x_{4}=0,x_{5}=0,x_{6}=0,x_{7}=0,x_{8}=r, \\ 
& x_{9}=0,x_{10}=s,x_{11}=0,x_{12}=-x_{10},x_{13}=u,x_{14}=0,x_{15}=-r+s, \\ 
& x_{16}=0,x_{17}=0,x_{18}=-r+s,x_{19}=0,x_{20}=0;\ r,s,u\in {\mathbb{R}}\}.%
\end{array}%
\end{equation*}%
This shows that, at $\mathbf{q}_{0}$, the linkage can perform 5-dimensional
first-order motions, so that the differential DOF is $\delta _{\mathrm{diff}%
}\left( \mathbf{q}_{0}\right) =\dim K_{\mathbf{q}_{0}}^{1}V=5$. The
higher-order cones are $K_{\mathbf{q}_{0}}^{i}=K_{\mathbf{q}_{0}}^{2},i\geq
3 $, and thus $C_{\mathbf{q}_{0}}^{\mathrm{K}}V=K_{\mathbf{q}_{0}}^{2}$.
This shows that there is exactly one manifold passing through $\mathbf{q}_{0}
$. Consequently, the linkage admits well-defined 3-dimensional smooth
motions, but no bifurcation, and the local finite DOF has a lower bound $%
\delta _{\mathrm{loc}}\left( \mathbf{q}_{0}\right) \geq \dim C_{\mathbf{q}%
_{0}}^{\mathrm{K}}V=3$. The DOF could be higher than 3 if $\mathbf{q}_{0}$
belongs to a subvariety of $V$ (with higher dimension ) which does not allow
smooth motions, e.g. cusps. The latter is not captured by the kinematic
tangent cone (or any analysis based on higher-order derivatives of
constraints). This is addressed in section \ref{secAproxV}.

\section{Differential Mobility at Smooth Finite Motions%
\label{secDiffMob}%
}

The analysis of the local finite mobility does not reveal the differential
mobility when passing through $\mathbf{q}_{0}$. It remains to determine i)
the differential DOF when the linkage performs finite motions through $%
\mathbf{q}_{0}$, and ii) whether the differential DOF is locally constant in
a neighborhood of $\mathbf{q}_{0}$ in $V$. The first allows to conclude
whether the linkage is shaky. The latter is crucial in order to assert
whether $\mathbf{q}_{0}$ is a kinematic singularity. It was recently
recognized (although this phenomenon has been know for a long time already)
that a mechanism can exhibit kinematic singularities even when the c-space
is a smooth manifold \cite{LeeMueller-IDETC2019}.

Denote with $\mathbf{J}_{%
\bm{\alpha}%
\bm{\beta}%
}$ the $k\times k$ submatrix of $\mathbf{J}$, containing elements $J_{ij}$
according to the index sets $%
\bm{\alpha}%
=\{\alpha _{1},\ldots ,\alpha _{k}\},\alpha _{i-1}<\alpha _{i}$ and $%
\bm{\beta}%
=\{\beta _{1},\ldots ,\beta _{k}\},\beta _{j-1}<\beta _{j}$. The $%
\bm{\alpha}%
\bm{\beta}%
$-minor of $\mathbf{J}$ of order $k$ is then $m_{%
\bm{\alpha}%
\bm{\beta}%
}(\mathbf{q}):=\det \mathbf{J}_{%
\bm{\alpha}%
\bm{\beta}%
}(\mathbf{q})$. Its $i$th derivative is denoted with $M_{%
\bm{\alpha}%
\bm{\beta}%
}^{\left( i\right) }%
\hspace{-0.4ex}%
(\mathbf{q},\dot{\mathbf{q}},\ldots ,\mathbf{q}^{\left( i\right) }){:=}\frac{%
d^{i}}{dt^{i}}m_{%
\bm{\alpha}%
\bm{\beta}%
}(\mathbf{q})$. The set of points where the constraint Jacobian has rank
less than $k$ can be defined in terms of the $k$-minors \cite%
{JMR2018,CISMMueller2019}%
\begin{equation}
\begin{array}{ll}
L_{k}\,:{=}\left\{ \mathbf{q}\in {\mathbb{V}}^{20}|\right. & f_{l}\left( 
\mathbf{q}\right) =\mathbf{I},m_{%
\bm{\alpha}%
\bm{\beta}%
}(\mathbf{q})=0,l=1,2,3 \\ 
& \forall \left\vert 
\bm{\alpha}%
\right\vert 
\hspace{-0.5ex}%
=%
\hspace{-0.5ex}%
\left\vert 
\bm{\beta}%
\right\vert 
\hspace{-0.5ex}%
=%
\hspace{-0.3ex}%
k,%
\bm{\alpha}%
\subseteq \{1,\ldots ,6\},%
\bm{\beta}%
\subseteq \{1,\ldots ,n\}%
\big%
\}.%
\end{array}
\label{Rk2}
\end{equation}%
Of interest are the finite smooth motions with rank less than $k$. Tangents
to such motions through $\mathbf{q}\in V$ form the kinematic tangent cone to 
$L_{k}$, determined by the sequence%
\begin{equation}
{C_{\mathbf{q}}^{\text{K}}}L_{k}=K_{\mathbf{q}}^{k,\kappa }\subset \ldots
\subset K_{\mathbf{q}}^{k,3}\subset K_{\mathbf{q}}^{k,2}\subset {K_{\mathbf{q%
}}^{k,1}}  \label{CqLk}
\end{equation}%
with the $i$th-order cone defined by the $i$th-order derivatives of
constraints and minors 
\begin{equation}
\begin{array}{ll}
K_{\mathbf{q}}^{k,i}:=\{\mathbf{x}|\exists \mathbf{y},\mathbf{z},\ldots \in {%
\mathbb{R}}^{20}: & H_{l}^{\left( 1\right) }%
\hspace{-0.6ex}%
\left( \mathbf{q},\mathbf{x}\right) =H_{l}^{\left( 2\right) }%
\hspace{-0.6ex}%
\left( \mathbf{q},\mathbf{x},\mathbf{y}\right) =\ldots \mathbf{=}%
H_{l}^{\left( i\right) }%
\hspace{-0.6ex}%
\left( \mathbf{q},\mathbf{x},\mathbf{y},\mathbf{z,\ldots }\right) =\mathbf{0}%
, \\ 
& M_{%
\bm{\alpha}%
\bm{\beta}%
}^{\left( 1\right) }%
\hspace{-0.6ex}%
\left( \mathbf{q},\mathbf{x}\right) =M_{%
\bm{\alpha}%
\bm{\beta}%
}^{\left( 2\right) }%
\hspace{-0.6ex}%
\left( \mathbf{q},\mathbf{x},\mathbf{y}\right) =\ldots =M_{%
\bm{\alpha}%
\bm{\beta}%
}^{\left( i\right) }%
\hspace{-0.6ex}%
\left( \mathbf{q},\mathbf{x},\mathbf{y},\mathbf{z,\ldots }\right) =0, \\ 
& \forall 
\bm{\alpha}%
\subseteq \{1,\ldots ,6\},%
\bm{\beta}%
\subseteq \{1,\ldots ,n\},\left\vert 
\bm{\alpha}%
\right\vert 
\hspace{-0.5ex}%
=%
\hspace{-0.5ex}%
\left\vert 
\bm{\beta}%
\right\vert 
\hspace{-0.5ex}%
=k,l=1,2,3\}.%
\end{array}
\label{Kki}
\end{equation}%
Calculation of the derivatives of the minors is again possible efficiently
by simple vector operations \cite{Mueller-MMT2019}, also available as
Mathematica$^{\copyright }$ implementation \cite{Mueller-MendeleyData2019}.

In the reference configuration $\mathbf{q}_{0}$, the Jacobian has $\mathrm{%
rank}~\mathbf{J}\left( \mathbf{q}_{0}\right) =15$, while it can not have
rank higher than $3\cdot 6=18$. Thus $L_{k},k=16,17,18$ must be
investigated. The computation yields that all derivatives of the minors of
order $k=16,17,18$, $M_{%
\bm{\alpha}%
\bm{\beta}%
}^{\left( i\right) }=0,\left\vert 
\bm{\alpha}%
\right\vert 
\hspace{-0.5ex}%
=%
\hspace{-0.5ex}%
\left\vert 
\bm{\beta}%
\right\vert 
\hspace{-0.5ex}%
=16,17,18$ are zero for solutions of (\ref{ConstrHi}). It is concluded that
the Jacobian has locally constant rank 15 for any smooth motion through $%
\mathbf{q}_{0}$. Thus the linkage has a locally constant differential DOF $%
\delta _{\mathrm{diff}}\left( \mathbf{q}\right) =n-\mathrm{rank}~\mathbf{J}%
\left( \mathbf{q}\right) =20-15=5,\mathbf{q}\in U\left( \mathbf{q}%
_{0}\right) \cap V$. Since its differential and local DOF are permanently
different, 
\mbox{the linkage is shaky of degree $\delta
_{\mathrm{diff}}-\delta _{\mathrm{loc}}=2$.}

\section{Local Approximation of the Configuration Space%
\label{secAproxV}%
}

The above analysis does not capture singularities where no smooth motions
are possible, and the reference configuration $\mathbf{q}_{0}$ may still be
a singularity where $V$ is not locally the intersection of manifolds. This
happens at a dead-point/motion-reversal point (also called stationary
singularity). At such configurations the c-space is not the intersection of
manifolds so that no tangent can be defined at that point, e.g. at cusps 
\cite{CISMMueller2019,LopezCustodio-MMT2019}. This can be checked by means
of a local approximation of the c-space $V$ using the $k$th-order
Taylor-series expansion of the constraint mappings at $\mathbf{q}$ 
\begin{equation}
f_{l}\left( \mathbf{q}+\mathbf{x}\right) =f_{l}\left( \mathbf{q}\right)
+\sum_{k\geq 1}\frac{1}{k!}\mathrm{d}^{k}f_{l,\mathbf{q}}\left( \mathbf{x}%
\right) .  \label{fTaylor}
\end{equation}%
Since $f_{l}\left( \mathbf{q}\right) =\mathbf{I}$, for $\mathbf{q}\in V$,
the $k$th-order approximation of $V$ at $\mathbf{q}$ is given by%
\begin{equation}
V_{\mathbf{q}}^{k}:=\{\mathbf{x}\in {\mathbb{R}}^{n}|\mathrm{d}f_{l,\mathbf{q%
}}\left( \mathbf{x}\right) +\frac{1}{2}\mathrm{d}^{2}f_{l,\mathbf{q}}\left( 
\mathbf{x}\right) +\ldots +\frac{1}{k!}\mathrm{d}^{k}f_{l,\mathbf{q}}\left( 
\mathbf{x}\right) =\mathbf{0},l=1,\ldots ,3\}.  \label{Vk}
\end{equation}%
An efficient recursive and explicit expression of the differentials $\mathrm{%
d}^{k}f_{l,\mathbf{q}}$ was reported in \cite%
{Mueller-MMT2019,CISMMueller2019}, and a Mathematica$^{\copyright }$
implementation can be found in \cite{Mueller-MendeleyData2019}. The
computation yields $V_{\mathbf{q}_{0}}^{1}={K_{\mathbf{q}_{0}}^{1}}$ and $V_{%
\mathbf{q}_{0}}^{2}={K_{\mathbf{q}_{0}}^{2}}$, which confirms that the
linkage can only perform the smooth 3-dimensional motions.

\section{Conclusion}

It was shown that, in the reference configuration, the three-loop linkage
presented by Wohlhart \cite{Wohlhart-ARK2004} has finite local DOF 3, a
locally constant differential DOF 5, and that the reference configuration is
regular. Consequently, the linkage can perform finite 3-dimensional smooth
motions. Since the differential DOF exceeds the local DOF at all regular
points, it is shaky of degree 2.

The linkage is also overconstrained. The generic (topological) DOF (of any
spatial linkage with the given topology) is given by the
Chebychev-Kutzbach-Gr\"{u}bler formula, which can be expressed as $\delta _{%
\mathrm{top}}=\sum_{i\in J}\delta _{i}-6\gamma =n-6\gamma $. Applied to this
linkage, the generic DOF is $\delta _{\mathrm{top}}=20-6\cdot 3=2$, so that
it is overconstrained of degree 1.

In conclusion, the reference configuration is a constraint singularity but
not a kinematic singularity or a c-space singularity. It should be remarked
that the three-loop linkage shown in Fig. 4 of \cite{FAYET1995_1} (repeated
as Fig. 3 in \cite{Wohlhart-ARK2004}) shows similar interesting properties.
The presented analysis only reveals the mobility at the reference
configuration (which is arbitrary though). A global analysis will have to
resort to an algebraic formulation and use tools from geometric algebra.

\begin{acknowledgement}
This work has been support by the LCM K2 Center for Symbiotic Mechatronics
within the framework of the Austrian COMET-K2 program
\end{acknowledgement}

\bibliographystyle{spmpsci}
\bibliography{Fayet-Wolhart_ARK2020}

\begin{thebibliography}{10}
\providecommand{\url}[1]{{#1}}
\providecommand{\urlprefix}{URL }
\expandafter\ifx\csname urlstyle\endcsname\relax
  \providecommand{\doi}[1]{DOI~\discretionary{}{}{}#1}\else
  \providecommand{\doi}{DOI~\discretionary{}{}{}\begingroup
  \urlstyle{rm}\Url}\fi

\bibitem{Baker1980}
Baker, J.E.: On relative freedom between links in kinematic chains with
  cross-jointing.
\newblock Mechanism and Machine Theory \textbf{15}(5), 397 -- 413 (1980)

\bibitem{Davies1981}
Davies, T.: Kirchhoff's circulation law applied to multi-loop kinematic chains.
\newblock Mechanism and Machine Theory \textbf{16}(3), 171 -- 183 (1981)

\bibitem{Rico-ARK2006}
Diez-Mart{\'i}nez, C.R., Rico, J.M., Cervantes-S{\'a}nchez, J.J., Gallardo, J.:
  Mobility and connectivity in multiloop linkages.
\newblock In: J.~Lennar{\v{c}}i{\v{c}}, B.~Roth (eds.) Advances in Robot
  Kinematics, pp. 455--464. Springer Netherlands (2006)

\bibitem{FAYET1995_1}
Fayet, M.: M\'{e}canismes multi-boucles—i détermination des espaces de
  torseurs cinématiques dans un mécanisme multi-boucles quelconque.
\newblock Mech. Mach. Theory \textbf{30}(2), 201 -- 217 (1995)

\bibitem{Lerbet1999}
Lerbet, J.: Analytic geometry and singularities of mechanisms.
\newblock ZAMM, Z. angew. Math. Mech. \textbf{78(10b)}, 687--694 (1999)

\bibitem{LopezCustodio-MMT2020}
Lopez-Custodio, P., M{\"u}ller, A., Kang, X., Dai, J.: Tangential intersection
  of branches of motion.
\newblock Mech. Mach. Theory \textbf{147} (2020)

\bibitem{LopezCustodio-MMT2019}
Lopez-Custodio, P., M{\"u}ller, A., Rico, J., Dai, J.: A synthesis method for
  1-dof mechanisms with a cusp in the configuration space.
\newblock Mech. Mach. Theory \textbf{132}, 154--175 (2019)

\bibitem{JMR2016}
M\"{u}ller, A.: Local kinematic analysis of closed-loop linkages -mobility,
  singularities, and shakiness.
\newblock ASME J. of Mech. and Rob. \textbf{8} (2016)

\bibitem{JMR2018}
M\"{u}ller, A.: Higher-order analysis of kinematic singularities of lower pair
  linkages and serial manipulators.
\newblock ASME J. Mech. Rob. \textbf{10(1)} (2018)

\bibitem{Robotica2018}
M\"{u}ller, A.: Topology, kinematics, and constraints of multi-loop linkages.
\newblock Robotica \textbf{36(11)}, 1641--1663 (2018)

\bibitem{Mueller-MendeleyData2019}
M{\"u}ller, A.: Data for: An overview of formulae for the higher-order
  kinematics of lower-pair chains with applications in robotics and mechanism
  theory.
\newblock Mendeley Data, v1  (2019)

\bibitem{CISMMueller2019}
M\"{u}ller, A.: Local investigation of mobility and singularities of linkages.
\newblock In: A.~M\"{u}ller, D.~Zlatanov (eds.) Singular Configurations of
  Mechanisms and Manipulators, CISM 589. Springer (2019)

\bibitem{Mueller-MMT2019}
M{\"u}ller, A.: An overview of formulae for the higher-order kinematics of
  lower-pair chains with applications in robotics and mechanism theory.
\newblock Mech. Mach. Theory \textbf{142} (2019)

\bibitem{LeeMueller-IDETC2019}
M{\"u}ller, A., Li, Z.: Mechanism singularities revisited from an algebraic
  viewpoint.
\newblock In: 43rd Mechanisms and Robotics Conference (MR) / ASME International
  Design Engineering Technical Conferences (IDETC), Anaheim, CA, USA (August
  18-21, 2019)

\bibitem{SeligBook}
Selig, J.: Geometric Fundamentals of Robotics.
\newblock Springer (2005)

\bibitem{Wohlhart-ARK2004}
Wohlhart, K.: Screw spaces and connectivities in multiloop linkages.
\newblock In: J.~Lenar{\v{c}}i{\v{c}}, C.~Galletti (eds.) On Advances in Robot
  Kinematics, pp. 97--104. Springer Netherlands (2004)

\end{thebibliography}

\end{document}